\documentclass[letterpaper, 10pt, conference]{ieeeconf}      

\IEEEoverridecommandlockouts                              

\usepackage[usenames, dvipsnames]{color}
\usepackage[noadjust]{cite}
\usepackage{tabularx}
\usepackage{array}
\usepackage{booktabs}
\usepackage{graphicx}
\usepackage{caption}
\usepackage{subcaption}
\usepackage{float}
\usepackage{amsmath}
\usepackage{algorithmicx}
\usepackage{subfiles}
\usepackage{multirow}
\usepackage{mathrsfs}
\usepackage{blindtext}
\usepackage{amsfonts}
\graphicspath{{images}{../images/}}
\usepackage[mathcal]{euscript}

\usepackage[usenames,dvipsnames,table]{xcolor}
\usepackage{soul}
\usepackage[export]{adjustbox}

\title{\LARGE \bf
GSGFormer: Generative Social Graph Transformer for Multimodal Pedestrian Trajectory Prediction 
}

\author{Zhongchang Luo$^{1, 2}$\thanks{$^{1}$~Ecole des Ponts, Marne-la-Vallée, France, email: FirstName.LastName@eleves.enpc.fr}\thanks{$^{2}$~Akkodis Research, 78280 Guyancourt, France, email: FirstName.LastName@akkodis.com}, Marion Robin$^{2}$ and Pavan Vasishta$^{2}$}

\begin{document}

\maketitle
\thispagestyle{empty}
\pagestyle{empty}
\setlength{\belowdisplayskip}{2pt}
\widowpenalty10000
\clubpenalty10000
\addtolength{\abovedisplayskip}{-5pt}

\begin{abstract}
 
Pedestrian trajectory prediction, vital for self-driving cars and socially-aware robots,
is complicated due to intricate interactions between pedestrians, their environment, and other Vulnerable Road Users.
This paper presents GSGFormer, an innovative generative model adept at predicting pedestrian trajectories by considering these complex interactions and offering a plethora of potential modal behaviors. 
We incorporate a heterogeneous graph neural network to capture interactions between pedestrians, semantic maps, and potential destinations. 
The Transformer module extracts temporal features, while our novel CVAE-Residual-GMM module promotes diverse behavioral modality generation. 
Through evaluations on multiple public datasets, GSGFormer not only outperforms leading methods with ample data but also remains competitive when data is limited.


\end{abstract}


\section{Introduction} \label{sec:introduction}

Connected mobility is changing all our lives every day, be it in the field of social robotics or self driving cars. Specifically in the case of self driving cars, the importance of predicting Vulnerable Road Users' (VRUs) positions becomes paramount. It also becomes a hard problem to solve since most VRU motion, especially that of pedestrians and bicyclists on the street are not deterministic but highly stochastic. Cars, to navigate safely around such VRUs, need to have a high degree of Situational Awareness (SA). In literature, one of the methods exploited to achieve this high level of SA is to use the Social Force model \cite{helbing1995social}. Other works have dealt with the importance of the semantics and the inherent goals existing in the environment in establishing pedestrian behaviour \cite{vasishta2017natural,vasishta2018building}. 

In this work, we aim to model pedestrian behaviour in urban areas so as to create a risk map of the environment that can then be used by other parts of the AV system to navigate through such an area. We do this by modelling the social forces around the car in the environment, identifying the different possible goals for each pedestrian in the scene and utilising the map information around each detected pedestrian at each time step in their trajectory.

In summary, our contributions presented in this work are as follows:
(1) we propose a novel model architecture that captures social and environmental interactions to accurately model spatio-temporal pedestrian trajectories over a short to medium term horizon (2) this architecture incorporates a novel goal identification module that generates multiple likely end points for each observed pedestrian partial trajectory. (3) A generative model that beats the state of the art on two drone datasets proving that the model can generalise to similar environmental scenarios. 

Section \ref{sec:related_work} discusses existing work in the domain of multi-modal,  medium-horizon, trajectory prediction for VRUs. Section \ref{sec:approach} discusses our theoretical approach and presents the CVAE based model for generating multi-modal pedestrian trajectories. Section \ref{sec:implementation} provides a detailed explanation on the implementation and results of our approach on various open datasets as well as ablation studies. 
\section{Related Work} \label{sec:related_work}

\subsection{Sequence modelling}

Capturing temporal sequences and predicting future states have long been studied, using classical methods or neural network methods, especially in the field of pedestrian trajectory prediction. In the former category, some methods use Hidden Markov models and self organising graphs to predict trajectories \cite{vasishta2018building}, while in the latter category Recurrent Neural Networks have been largely popular in capturing pedestrian behaviour \cite{alahi2016social, salzmann2020trajectron++, huang2019stgat, gupta2018social, lee2017desire, vemula2018social}. In other cases, CNNs like in  Time-Extrapolator Convolution Neural Network (TXP-CNN) \cite{mohamed2020social} and U-Net \cite{mangalam2021goals} have also been used to extract temporal features from pedestrian trajectories.

Recently however, the sequence Transformer which was originally proposed in the NLP domain \cite{vaswani2017attention}, has proven to be very successful in modelling pedestrian behaviour \cite{giuliari2021transformer,yuan2021agentformer,yu2020spatio, zhang2023forceformer}. This class of architectures have come in handy when used in scenarios with partial observability and partial trajectories, which are difficult for RNNs to deal with. 

\subsection{Social forces modelling}
The concept of social forces in humans was first propounded in \cite{helbing1995social} which has then been applied to trajectory forecasting \cite{vasishta2017natural}. Social forces are those forces that act on an ego agent in a built environment as a function of its surroundings and other agents present in the scene, causing an attractive or repulsive interactions with them. In brief, these can be commonly classified as neighbour interactions, map interactions and goal based interactions. Modelling of these forces using neural networks have been studied in mechanisms such as Social pooling \cite{alahi2016social, gupta2018social}, attention mechanisms \cite{vemula2018social, sadeghian2019sophie, xu2022socialvae} and graph-based neural networks for homogeneous \cite{huang2019stgat, yu2020spatio, xu2023eqmotion} and heterogeneous interactions \cite{mo2022multi,chiara2022goal}. 

Map-derived interactions have also been studied by embedding the map information as an aid to prediction \cite{salzmann2020trajectron++, yuan2021agentformer, messaoud2021trajectory}while \cite{sadeghian2019sophie} utilizes physical attention to take in account the map information. This information in turn can be used to identify and predict goal based interactions \cite{mangalam2021goals,chiara2022goal} or to generate less expensive goal heatmaps \cite{mangalam2020not, wang2022stepwise, yue2022human}

\subsection{Multi-modal output}  
Considering that human motion is inherently stochastic, it becomes necessary to handle this uncertainty while performing prediction. This is handled in different ways in different works.  A simple random noise is added or concatenated to the input of the trajectory generator in some works \cite{huang2019stgat, yu2020spatio}. \cite{gupta2018social, sadeghian2019sophie} use GANs adding adversarial loss to generate prediction with higher quality. CVAE models have also been popular to implement multi-modal trajectory prediction \cite{lee2017desire, salzmann2020trajectron++, yuan2021agentformer, xu2022socialvae, yue2022human} due to its ability to learn the latent variable distribution from prior conditions. To explicitly express different modalities, certain anchor-conditioned methods have been proposed \cite{huang2023multimodal}, such as predicted endpoint conditioned (PEC) \cite{mangalam2020not} and prototype trajectory conditioned (PTC) \cite{kothari2021interpretable} frameworks to encourage more explainable predictions.

In our work, we take ideas from each of these systems such as using Transformers to capture temporal trajectories,  Graph Networks to capture varied and heterogeneous interactions and CVAEs to model the multi-modality of human behaviour. The exact method in which these components are assembled together in our architecture will be described in the next section. 
\section{Theoretical Approach} \label{sec:approach}

\subsection{Problem formulation}

Our problem of multimodally predicting pedestrian trajectories can be formally prescribed as follows. Starting with a bird's eye view (BEV) scene of an environment captured in RGB, semantically segmented into different classes, denoted by $\mathcal{M}$ contains a history of positions of each pedestrian in this scene. For each scene, there exist a finite number of goals that can be reached by a pedestrian denoted by $\mathcal{G}$. Each pedestrian also interacts with every other agent in the scene, denoted by $\mathcal{E}$. 

Given this information, the model aims to predict the distribution of pedestrian's trajectory for $H$ timesteps in the future.

We aim to predict the distribution of the future trajectory based on all the history information we have, and then generate multi-modal trajectories for each agent. We learn the parameters $\theta$ of the probability $P_\theta(\mathbf{Y}|\mathbf{X}, \mathcal{M}, \mathcal{G}, \mathcal{E} )$ with $\mathbf{X}$ denoting trajectory history and $\mathbf{Y}$ denoting the predicted trajectory. 

For an observed agent $i$, the model outputs the predicted future positions in the next $H$ time steps from the current time $t$, defined as $\mathbf{Y}_i=(\mathbf{Y}_i^{t+1}, \mathbf{Y}_i^{t+2},... , \mathbf{Y}_i^{t+H})$. $\mathbf{Y}_i^t=(x_i^t, y_i^t)$ is the position coordinate of the agent with ${x, y \in \mathbb{R}}$. The input of the model is the history information in the past $n$ time steps, which can be denoted as $\mathbf{X}_i=(\mathbf{X}_i^{t-n+1}, \mathbf{X}_i^{t-n+2},... , \mathbf{X}_i^{t})$. In order to consider the effect of social forces, the history states of ego agent and all neighbors as well as the semantic map information are input to the model. We assume that the number of agents in the area of the shared space around the ego agent (including the ego agent) is a time varying variable $N^t$. The input at each time step can be denoted as $\mathbf{X}_i^t=(\mathbf{S}_a^t, \mathbf{I}_i^t)_{a=1}^{N^t}$, where $\mathbf{S}_a^t=(x_a^t, y_a^t, \dot{x}_a^t, \dot{y}_a^t)$ represents the state of the agent $a$, and $\mathbf{I}_i^t$ denotes the semantic map information at time $t$ around the ego agent.

\subsection{Proposed model}
\begin{figure}[!ht]
  \centering
  \includegraphics[width=0.5\textwidth]{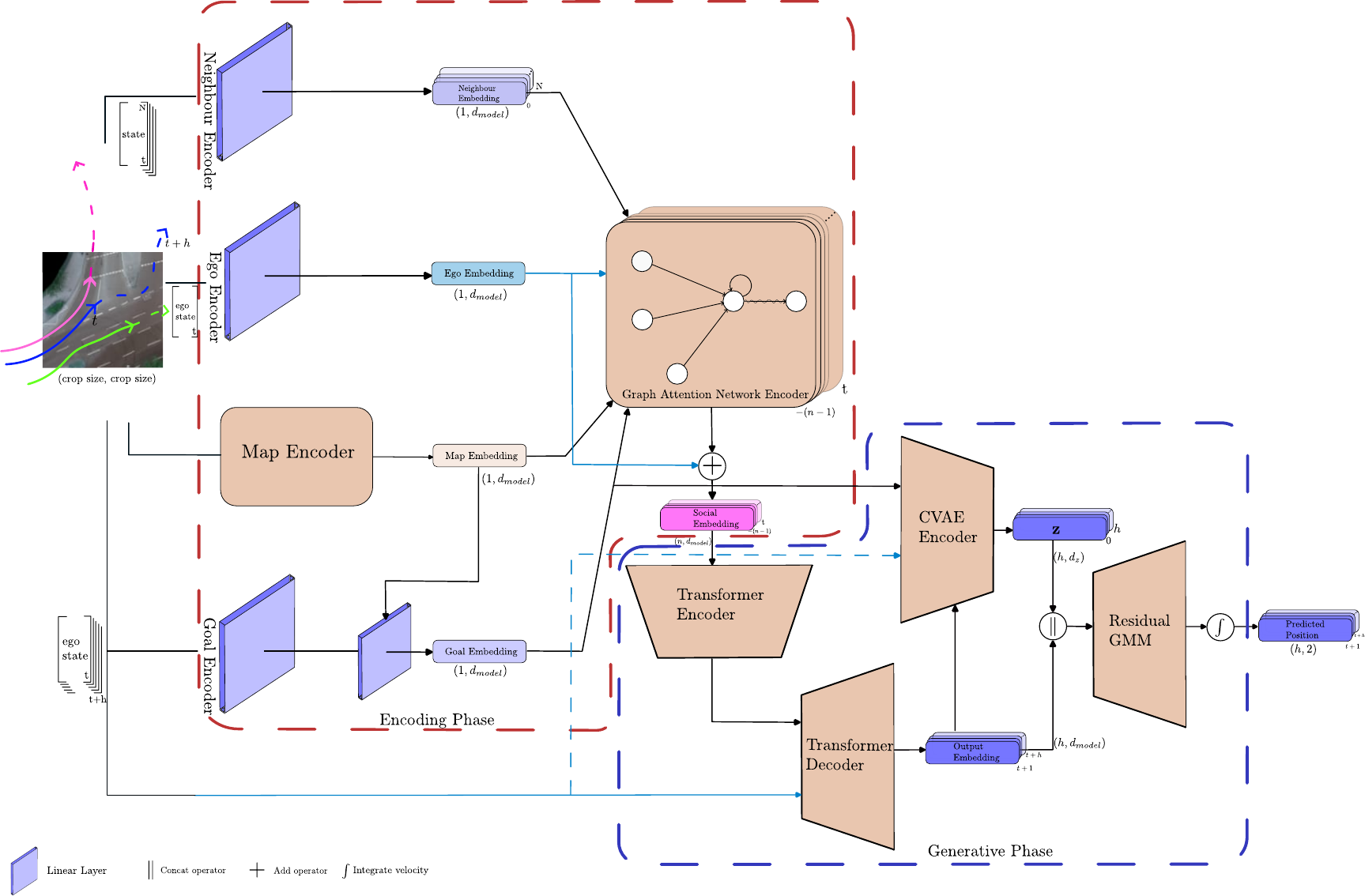}
  \caption{GSGFormer model overview. Consider various VRUs in a structured urban environment, in BEV as shown. Our model uses the understanding of the interplay of social forces between different VRUs and the environment in which the motion is captured. A map encoder captures the influence of the environment on the VRU under consideration while the other encoders - for encoding the ego motion of the VRU, identifying the different goals of this VRU and those of its neighbours - are fed into a Graph Attention Network Encoder to capture a ``Social Embedding'' within the environment. Social Embeddings for each time step are then passed on to the Generative Phase of the model, where different contexts from the Transformer Decoder and the CVAE Encoder are concatenated along the last dimension of the tensor and passed to the Residual GMM generator. Finally, positions for a time interval $h$ from current time $t$ are predicted by integrating these velocities over the time interval $dt$. Best viewed zoomed in.}
  \label{fig:fullnetwork}
\end{figure}

Our proposed model can be split into three main components, in two phases as depicted in Fig. \ref{fig:fullnetwork}. In the Encoding Phase, we encode the ego agent's interactions with the environment and other agents in the scene to create ``Social Embeddings''. These Embeddings are used along with other data to gererate multi-modal trajectories in the Generative Phase of the model. The three main components are the Graph Attention Encoder (GAE), the Temporal Transfmer Encoder-Decoder (TT) and the generative CVAE model for the ultimate output of multi-modal trajectories. Each of these components and their sub-components are described in detail below. 

\subsubsection{Agent State Encoders}
This submodule acts as input to the GAE. Each agent in the scene is classified as one of four classes -  pedestrian, bicycle, car and bus. The state of each class of agent is embedded by a corresponding linear layer. For each ego agent trajectory, neighbour tracks are identified by converting to an ego-centric coordinate, their position relative to the ego agent \cite{jia2022hdgt}, also embedded using a linear layer. All outputs of the corresponding linear layers are denoted by $d_{model}$.

\subsubsection{Map Encoder}\label{subsec:mapenc}

\begin{figure}[!ht]
  \centering
  \includegraphics[width=0.5\textwidth]{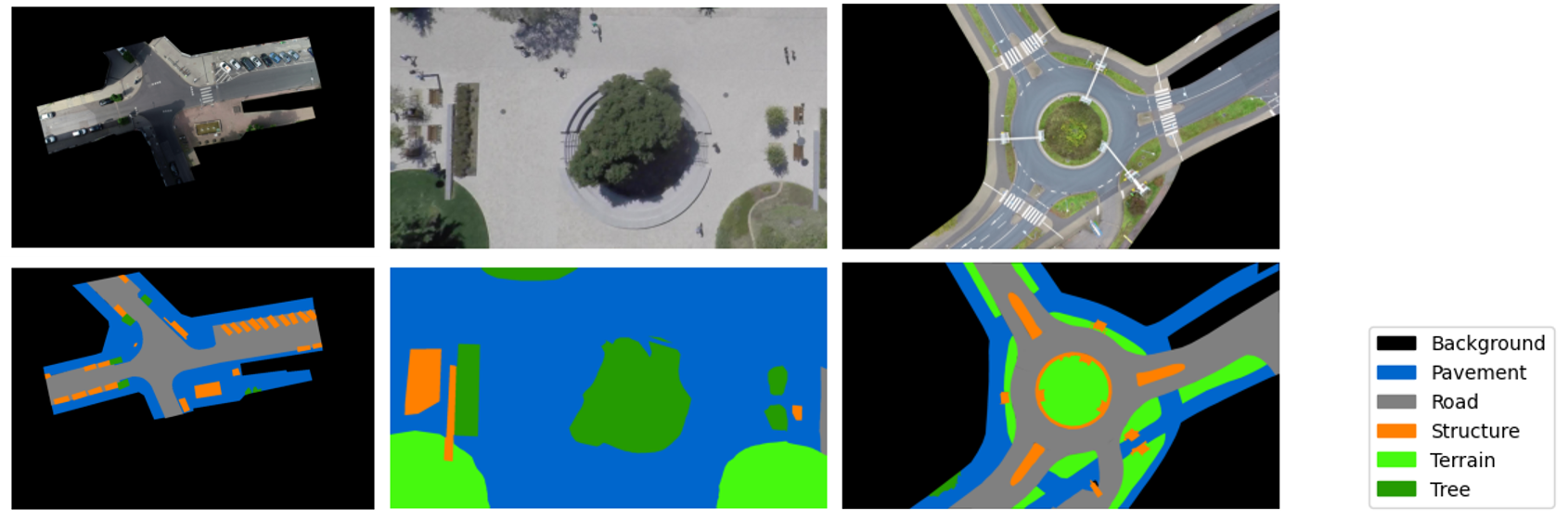}
  \caption{Examples from the inD, SDD and OpenDD dataset. Above is the original image and below is the segmentation mask.}
  \label{fig:masks}
\end{figure}

Encoding the Map first requires processing the RGB image a semantically segmented image, creating an abstraction over the real observation. To encode this, we use a U-Net \cite{ronneberger2015unet} architecture for semantically segmenting the RGB image. The semantic map labels are drawn from \cite{mangalam2021goals}. Where such labels are not available for training, they were manually annotated in 6 classes. In addition to the SDD\cite{robicquet2016learning} and inD\cite{inDdataset} datasets in the work\cite{mangalam2021goals}, we also used map images and our own segmentation annotations from the OpenDD dataset\cite{breuer2020opendd} as training set. The model was trained with Dice Loss \cite{sudre2017generalised} and the resulting segmentation masks are as shown in Fig. \ref{fig:masks}.

This mask output containing semantic information of the scene is then encoded using a 2D CNN to generate the map embedding of size $d_{model}$

\subsubsection{Goal Encoder}
This subcomponent is one of our novel contributions in this paper. We propose a 1D CNN to estimate the goal position. Initially, The state of the ego agent for each history time step is embedded using the agent state encoder. Then the embedding is stacked over the time dimension, resulting in a tensor of shape $(n \times d_{model})$. The 1D convolutional neural network extracts the temporal feature between the $n$ channels(time steps). The map embedding is concatenated to the output of the goal encoder and passed to a linear layer. The output of the network is a vector of size $d_{model}$. The vector then serves as the input condition for the CVAE-Residual-GMM module detailed further in \ref{subsec:cvaegmm}, facilitating the generation of a range of potential goals.

\subsubsection{Social Graph Attention Neural Network}

To model the social force acting on the ego agent, we introduce a heterogeneous graph representation. This representation comprises four distinct nodes: the ego agent, neighboring agents, the semantic map, and the goal populated by their respective embeddings. Physically, we only consider the effect of these embeddings unidirectionally on the ego agent at time step $t$. Then a heterogeneous graph attention network\cite{wang2019heterogeneous} performs both node-level and semantic-level attention aggregating the social information to the ego agent node. The output is the social force embedding in size of $d_{model}$ is embedded from the aggregated ego agent node. This embedding is then element-wise combined with the state embedding of the ego agent, producing the social embedding at time step $t$. 

\subsubsection{Temporal Transformer Model}  

Once the social information are encoded from the social graph of the ego agent at each time step, it is necessary to build the dependencies between graphs across different time steps. We employ a Transformer framework \cite{vaswani2017attention} to model the temporal relationship. The history sequence of social embedding are fed into the Transformer encoder as the source sequence with a fixed length n. Since the primary operation in Transformer is the attention mechanism which is not sensitive to the relative or the absolute position of the elements in the sequence, we add the "positional encodings" with the same dimension to each social embedding in the sequence. We compute positional encodings through sine and cosine functions, as described in \cite{vaswani2017attention}, which allows the model to easily learn to attend by relative positions. Meanwhile, it enables the model recognize any missing observations, as each comes with its distinct positional encoding.

The Transformer encoder processes the history sequence in parallel and outputs the memory providing the query for each observation time step. The Transformer decoder takes the embedded future ego agent's state sequence as the target sequence. The query of the memory is used to calculate the cross attention between the history sequence and future sequence. From the Transformer decoder, we obtain the predicted embedding sequence. Every embedding within this sequence serves as a condition for the waypoint CVAE-Residual-GMM Module (see in \ref{subsec:cvaegmm}), generating multi-modal states for each prediction horizon.

\subsubsection{CVAE-Residual-GMM Module}\label{subsec:cvaegmm} 

We propose a generative CVAE-Residual-GMM Module to realize the one-to-many mapping for both goal and waypoints. The module is composed of two main parts: a Conditional Variational Auto-Encoder(CVAE) \cite{sohn2015learning} and a Gaussian Mixture Model(GMM) \cite{dempster1977maximum}.  Let's denote the input condition as $\mathbf{X}$ and its corresponding output (be it goal position or waypoint state) as $\mathbf{Y}$. The high dimensional latent variable $\mathbf{Z}$ is drawn from the prior distribution $P_\theta(\mathbf{Z}|\mathbf{X})$, facilitated by a prior network. Throughout the training phase, given that the ground truth is accessible, $\mathbf{Z}$ is drawn from the posterior distribution $Q_\phi(\mathbf{Z}|\mathbf{X}, \mathbf{Y})$ by the recognition network. $\mathbf{Y}$ is finally generated from the distribution $P_\psi(\mathbf{Y}|\mathbf{X}, \mathbf{Z})$ through the generation network. 

Different from the vanilla CVAE, the generated $\mathbf{Y}$ is the sum of two parts: the output $\mathbf{\hat Y}$ from CVAE decoder decoding the concatenated $(\mathbf{X}, \mathbf{Z})$ and the residual $\mathbf{\Delta Y}$, which adheres to a GMM distribution. This distribution is parameterized by an MLP, defining means, variances, and component weights based on $(\mathbf{X}, \mathbf{Z})$. Notably, every waypoint within a single trajectory utilizes a consistent $\mathbf{Z}$ for each sampling iteration.

\subsubsection{Training Strategy}
During training, the most recent historical state, alongside the entire future state sequence, is concurrently fed into the Transformer decoder. A subsequent masking is applied to speed up the training and to ensure predictions for time step $t$ rely solely on preceding time steps. The latent variable $\mathbf{Z}$ is drawn both from $P_\theta(\mathbf{Z}|\mathbf{X})$ and $Q_\phi(\mathbf{Z}|\mathbf{X}, \mathbf{Y})$.

For inference, the decoder first receives the most recent historical state, followed by an auto-regressive procedure. The Transformer decoder's output embedding is then fed into the CVAE as condition $\mathbf{X}$. The latent variable $\mathbf{Z}$ is sampled from the prior network $P_\theta(\mathbf{Z}|\mathbf{X})$. The predicted state is calculated by summing up the output from CVAE decoder and the residual sampled from $P_\psi(\mathbf{\Delta Y}|\mathbf{X}, \mathbf{Z})$. This predicted state is subsequently appended in the time dimension to the previous input of the Transformer decoder to forecast the next time step. Notably, all waypoints in a single trajectory utilize a consistent $\mathbf{Z}$ throughout the sampling phase. 

The predicted position is determined by integrating the displacement of each time step, which is computed using velocity and time intervals. This method has been demonstrated to be more precise than directly utilizing the predicted position, as supported by \cite{salzmann2020trajectron++}.

\subsubsection{Loss function}

The loss function consists of two components: the waypoint and the goal loss, both represented through the CVAE loss format. The CVAE loss is expressed as the negative empirical lower bound (ELBO) \cite{sohn2015learning}. While the aim of CVAE is to maximize the ELBO, due to our reliance on gradient descent methods for optimizing the neural network, we aim to minimize the negative ELBO, detailed as follows:

\begin{equation}\label{eq:lossfunction}
\begin{aligned}
\mathcal{L} = \mathcal{L}_{\text {waypoint}} + \mathcal{L}_{\text {goal}} \\
\end{aligned}
\end{equation}

\begin{equation}\label{eq:ELBO}
\begin{aligned}
\mathcal{L}_{\text {CVAE}}= & -\mathbb{E}_{Q_\phi(\mathbf{Z}|\mathbf{Y}, \mathbf{X})}\left[\log P_\psi(\mathbf{Y}|\mathbf{X}, \mathbf{Z})\right]\\ &+\operatorname{KL}\left(Q_\phi(\mathbf{Z}|\mathbf{Y}, \mathbf{X}) \| P_\theta(\mathbf{Z}|\mathbf{X})\right)
\end{aligned}
\end{equation}

where $P_\theta(\mathbf{Z}|\mathbf{X})$ and $Q_\phi(\mathbf{Z}|\mathbf{Y}, \mathbf{X})$ are derived from the normal distribution probability determined by the mean and variance as parameterized by prior network and recognition network. $P_\psi(\mathbf{Y}|\mathbf{X}, \mathbf{Z})$ corresponds to the residual probability $P_\psi(\mathbf{\Delta Y}|\mathbf{X}, \mathbf{Z})$, which adopts a GMM distribution probability. This is because $\mathbf{Y}$ is the sum of $\mathbf{\hat{Y}}$ and $\mathbf{\Delta Y}$, where $\mathbf{\hat{Y}}$ is deterministic.

\section{Experiments and Results} \label{sec:implementation}

\subsection{Datasets} \label{subsec:datasets}
We train and evaluate our model on three stationary BEV datasets - the intersection Drone (\textbf{inD}) Dataset \cite{inDdataset}, the Stanford Drone Dataset (\textbf{SDD}) \cite{robicquet2016learning} and the \textbf{ETH/UCY} Dataset \cite{pellegrini2009you,lerner2007crowds}. 

\textbf{inD} dataset contains 32 drone recordings of various VRU trajectories of 5 different classes (pedestrian, bicycle, car, truck and bus) in 4 different scenes at German intersections. The raw data is samples in 25Hz. Training and test sets are split based on previous work done \cite{bertugli2021ac} such that scenes 1 and 2 of the dataset are considered as training data, scene 3 to be the validation set and scene 4 as the test set. 

\textbf{SDD} is a similar drone dataset, with 60 recordings at 8 unique scenes in the Stanford university campus, each scene containing a mixture of 6 classes of VRUs (pedestrians, skateboarders, bicyclists, cars, cars and buses). The raw data is sampled at 30 Hz. The entire dataset is split in two specific ways - one proposed in TrajNet \cite{sadeghian2018trajnet} and another full split as in DESIRE \cite{lee2017desire}. 

\textbf{ETH/UCY} is a classic dataset widely used for the evaluation for pedestrian trajectory prediction which contain only data of pedestrians with rich human-human interactions. The raw data is extracted and annotated at 2.5Hz. We follow the leave-one-out strategy as presented in previous works \cite{alahi2016social,gupta2018social} for evaluation. 

The first two datasets are more representative of scenarios that AVs can face in urban environments, especially in shared spaces than the classic ETH/UCY dataset. The background RGB frames in these datasets act as the maps for our model. In each dataset, we only concentrate on predicting trajectories of the pedestrian VRU class while utilizing the effects of the other VRU classes on these pedestrians. Another rationale for utilizing similar datasets is to investigate our model's robustness to different scenes and vantages.

\begin{table*}[!ht]
  \centering
  \small
  \begin{tabular}{p{3.3cm}|p{1.8cm}|p{1.8cm}|p{1.8cm}|p{1.8cm}|p{1.8cm}|p{1.8cm}}
      \hline \multicolumn{7}{c}{ Evaluation Metrics ($\textbf{minADE}_{20}$  / $\textbf{minFDE}_{20}$ )} \\
      \hline 
      \textbf{Method}& \textbf{ETH} & \textbf{HOTEL} & \textbf{UNIV} & \textbf{ZARA1} & \textbf{ZARA2} & \textbf{AVERAGE} \\
      \hline 
      Social-GAN \cite{gupta2018social} & 0.81/1.52 & 0.72/1.61 & 0.60/1.26 & 0.34/0.69 & 0.42/0.84 & 0.58/1.18 \\
      \hline 
      ST-GAT \cite{huang2019stgat} & 0.65/1.12 & 0.35/0.66 & 0.52/1.10 & 0.34/0.69 & 0.29/0.60 & 0.43/0.83 \\
      \hline 
      Y-Net \cite{mangalam2021goals} & 0.28/0.33 & 0.10/0.14 & 0.24/0.41 & 0.17/0.27 & 0.13/0.22 & 0.18/0.27 \\
      \hline 
      NSP-SFM \cite{yue2022human} & \textbf{0.25}/\textbf{0.24} & \textbf{0.09}/\textbf{0.13} & \textbf{0.21}/\textbf{0.38} & \textbf{0.16}/\textbf{0.27} & \textbf{0.12}/\textbf{0.20} & \textbf{0.17}/\textbf{0.24} \\
      \hline 
      GSGFormer(Ours) & 0.56/0.99 & 0.18/0.31 & 0.32/0.56 & 0.21/0.35 & 0.16/0.28 & 0.29/0.50 \\
      \hline
  \end{tabular}
  \caption{$\text{minADE}_{20}$ and $\text{minFDE}_{20}$ evaluated on ETH/UCY dataset in meters.}
  \label{table:eth/ucy}
\end{table*}

\subsection{Experimental Setup}
For each of the datasets chosen and all their corresponding scenes, we perform trajectory preprocessing as provided by OpenTraj tools \cite{amirian2020opentraj}. All VRU trajectories are sampled at 2.5Hz and velocities for these trajectories are calculated as $\Delta X |_{t}^{t+1} / \Delta t$ where $X$ represents the sampled position at time $t$ and $\Delta t$ is the sampling interval. Each observed partial trajectory lasts for 8 timesteps with a prediction horizon $h$ of 12 timesteps thus implying a medium term trajectory prediction window of $4.8s$ after an observation window of $3.2s$, consistent with the baselines (\ref{subsec:basemetric}) chosen to compare against our work. 

For each trajectory, a distance of $15m$ (or equivalent) is chosen as an effective threshold for neighbour interactions while a crop patch of $10m$ (or equivalent) is chosen for map interaction for social embeddings. Parentheses imply an estimation from pixel space to Cartesian space where direct measurements are unavailable in the dataset. 

The model is set up to be trained on a single NVIDIA RTX A6000 GPU with 48 GB VRAM. 

\subsection{Baselines and Metrics}\label{subsec:basemetric}
\paragraph{Baselines}
We compare our model's effectiveness with different baselines for each of the datasets described in Section \ref{subsec:datasets} since each dataset has its own merits and drawbacks. However, we explain some choices of baselines as below:

\textbf{Social GAN} \cite{gupta2018social}: One of the earliest generative architectures that focus on multi-modal trajectory prediction. They apply a an LSTM Encoder-Decoder architecture connected by a social pooling module which used to aggregate the social interactions between actors in the scene. 

\textbf{ST-GAT} \cite{huang2019stgat}: ST-GAT is a classic work on human trajectory prediction which uses graph attention network to model human interactions. LSTM is applied to model the temporal feature of the pedestrians.

\textbf{Y-Net} \cite{mangalam2021goals}: This multi-modal trajectory generation model uses goals, waypoints and segmented maps to predict pedestrian motion. However, it does not explicitly model any VRU interaction, thus acting as the perfect foil to compare our model against, especially in proving that social interactions form the backbone of any prediction model. 

\paragraph{Metrics} We use the Average Displacement Error (ADE) and Final Displacement Error (FDE) as metrics for measuring the efficacy of our prediction task. As we measure the output of a multi-modal generative model, we adopt a $\text{minADE}_{K}$ and $\text{minFDE}_{K}$ reporting strategy over a set of $K$ trajectories similar to prior approaches \cite{gupta2018social}.

\subsection{Implementation details}\label{subsec:implement}
All input states - ego state, neighbour states, goal states - are embedded in a space of $d_{model}=128$. We use a map patch centered on the ego agent parsed through a pretrained UNet architecture to extract semantic segments from the patch and subsequently embed it in a space of 128. The Graph Attention Network is composed of two input layers of size 128 in mean aggregation mode with 4 attention heads for each layer. The single layer Transformer encoder decoder module has a feed forward dimension set to 256 with 4 heads of multi-head attention. Each MLP in the CVAE encoder, decoder and GMM consists of two hidden layers of size $512\rightarrow256$ with ReLU activations on each.  The dimension of the latent variable is 16 for the goal and 32 for the waypoint. The number of mixture components is set as 4 for goal and 20 for waypoint.

The semantic segmentation network for the map encoder is a single model trained on both the drone datasets, with manual annotations in 6 classes where unavailable as presented in Sec. \ref{subsec:mapenc}. 

The model is trained using the Adam optimizer with a learning rate starting at 0.001 and decaying by a factor of 0.5 every 10 epochs for a total of 50 epochs with a batch size of 512. 

\subsection{Quantitative Results}
We organise our results by dataset, with the reported $\text{minADE}$ and $\text{minFDE}$ for 20 samples each. Clearly, for all datasets, lower scores are better for both metrics with the best result shown in bold. Table \ref{table:ind} compares metrics for the baselines with ours. As observed, our model beats the state of the art in one metric while performing at the state of the art in the other. 
\begin{table}[!ht]
    \centering
    \normalsize
    \begin{tabular}{c|cc}
        \hline 
        \textbf{Method} & $\textbf{minADE}_{20}$ & $\textbf{minFDE}_{20}$ \\
        \hline 
        Social-GAN \cite{gupta2018social} & 0.48 & 0.99 \\
        ST-GAT \cite{huang2019stgat} & 0.48 & 1.00 \\
        AC-VRNN \cite{bertugli2021ac} & 0.42 & 0.80 \\
        Y-net* \cite{mangalam2021goals} & 0.34 & 0.56 \\
        Goal-SAR \cite{chiara2022goal} & 0.31 & \textbf{0.54} \\
        GSGFormer (Ours) & \textbf{0.30} & 0.55 \\
        \hline
    \end{tabular}
    \caption{$\text{minADE}_{20}$ and $\text{minFDE}_{20}$ evaluated on the inD dataset with result presented in meters. * indicates the Y-Net \cite{mangalam2021goals} result is trained by \cite{chiara2022goal}.}
    \label{table:ind}
\end{table}

Table \ref{table:sdd} shows the compared metrics for two different splits for SDD - overlapping trajectories for the entire dataset at the top with the Trajnet split at the bottom. It can be seen that our model performs better than the state of the art when trained on the entire dataset compared to the Trajnet split. We conjecture that this is due to the Trajnet split not containing sufficient social interaction information among its trajectories.  

Table \ref{table:eth/ucy} compares metrics of the baselines with our approach, however as can be seen, our model does not perform as well on this dataset as the others. This is because of the relative simplicity of ETH/UCY scenes and the inability of the Map Encoder module of our approach to reach its full potential, proving our model performs best on large scale datasets with environmental diversity.

\begin{table}[!h]
    \centering
    \normalsize
    \begin{tabular}{c|cc}
        \hline 
        \textbf{Method} & $\textbf{minADE}_{20} $ & $\textbf{minFDE}_{20}$ \\
        \hline 
        Social-GAN \cite{gupta2018social} & 27.25 & 41.44 \\
        DESIRE \cite{lee2017desire} & 19.25 & 34.05 \\
        SimAug \cite{liang2020garden} & \textbf{10.27} & 19.71 \\
        P2T \cite{deo2020trajectory} & 10.97 & 18.40 \\
        GSGFormer (Ours) & \textbf{10.27} & \textbf{16.67} \\
        \hline 
        PECNet \cite{mangalam2020not} & 9.96 & 15.88 \\
        P2T \cite{deo2020trajectory} & 8.76 & 14.08 \\
        Y-Net \cite{mangalam2021goals} & 7.34 & 11.53 \\
        NSP-SFM \cite{yue2022human} & \textbf{6.52} & \textbf{10.61} \\
        GSGFormer (Ours) & 9.62 & 15.29 \\
        \hline
    \end{tabular}
    \caption{$\text{minADE}_{20}$ and $\text{minFDE}_{20}$ evaluated on the entire SDD dataset (above) and the TrajNet split (below), reported in pixels.}
    \label{table:sdd}
\end{table}

\subsection{Qualitative Results}
\begin{figure*}[!ht]
  \centering
  \begin{subfigure}[b]{0.20\textwidth}
    \begin{minipage}[b]{1.0\textwidth}
      \includegraphics[width=\textwidth]{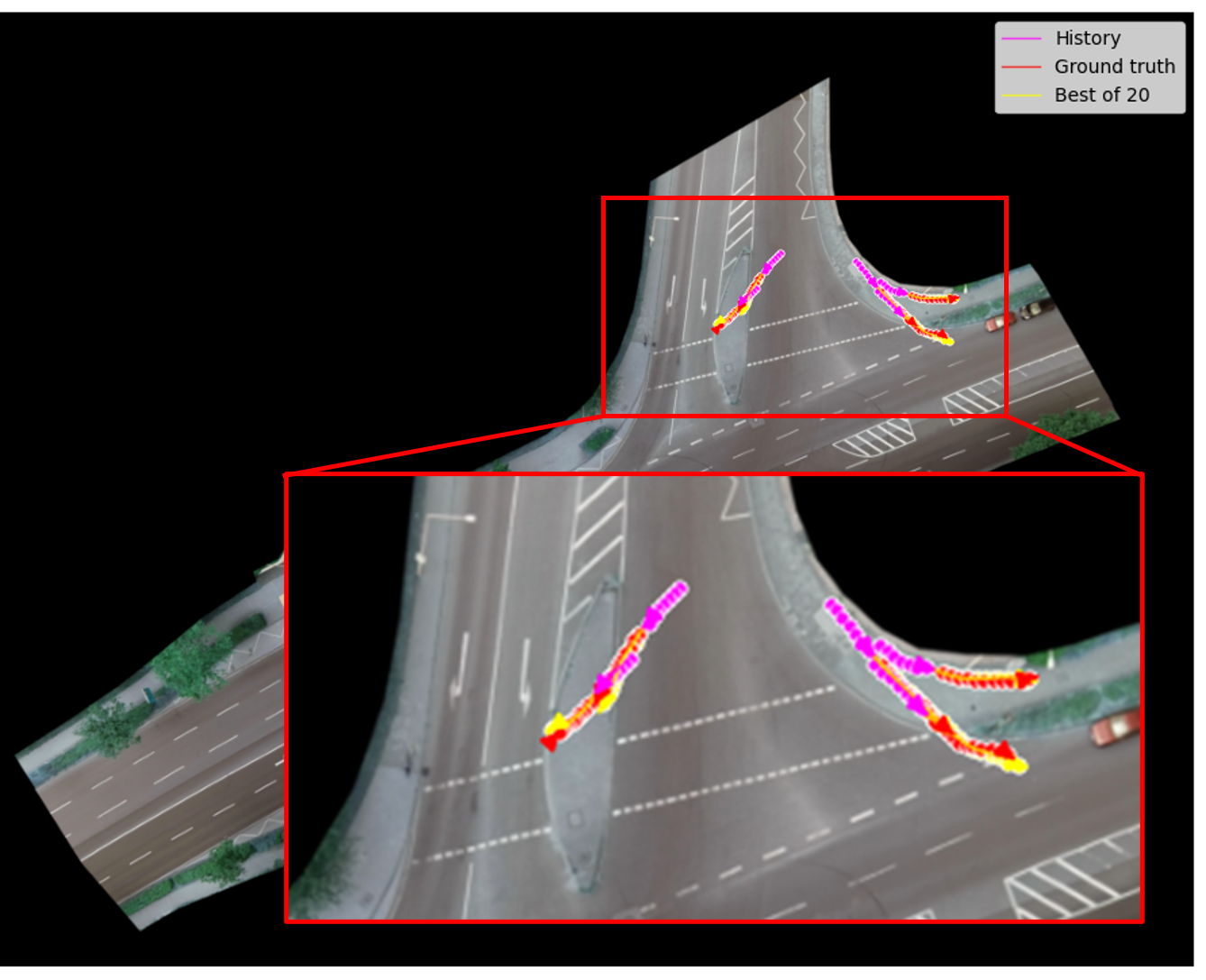}
      \caption{Best of 20 result (inD)}
      \label{fig:indbesttraj}
      \includegraphics[width=\textwidth]{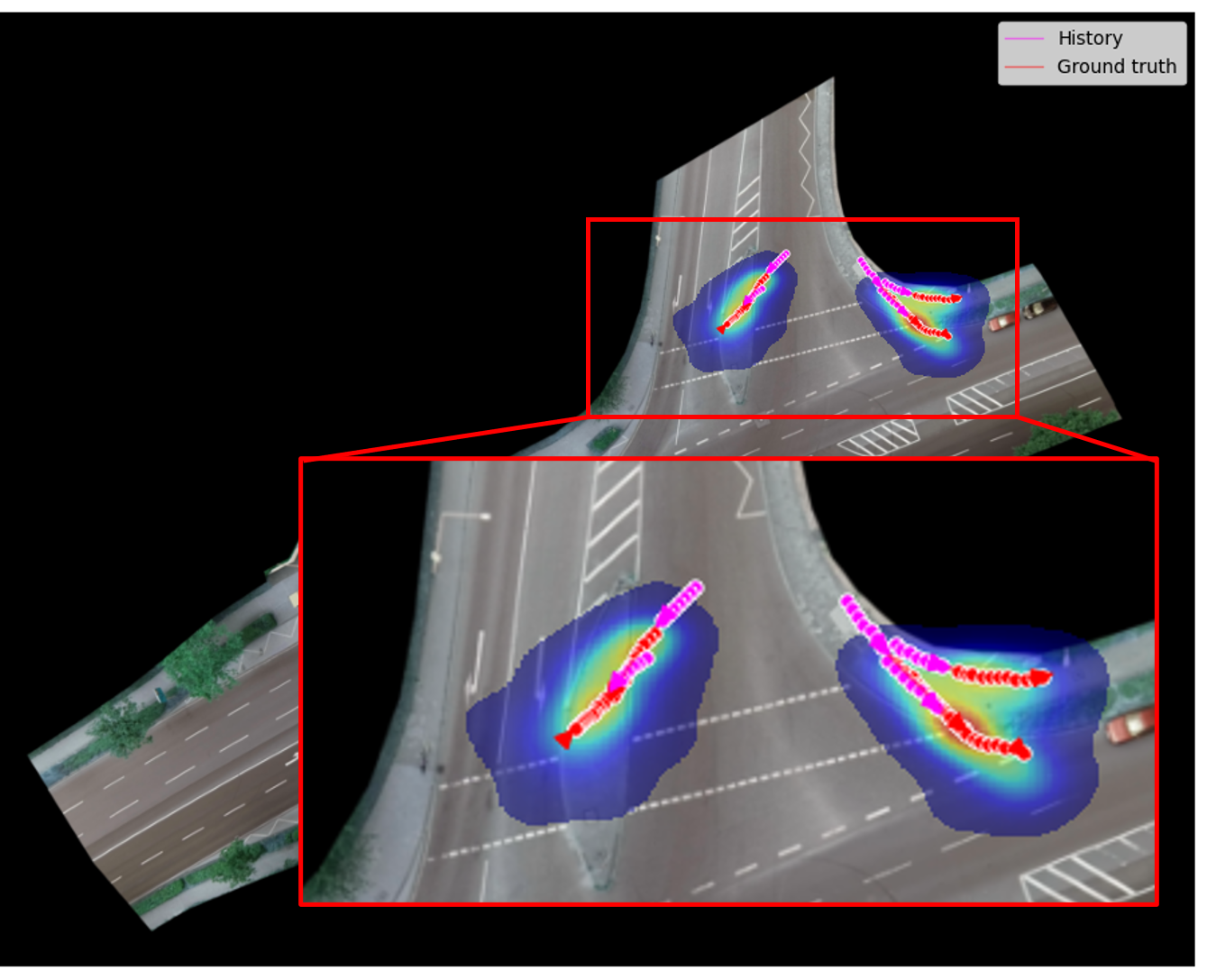}
      \caption{Risk map (inD)}
      \label{fig:indriskmap}
    \end{minipage}
  \end{subfigure}
  \begin{subfigure}[b]{0.25\textwidth}
    \centering
    \includegraphics[width=\textwidth]{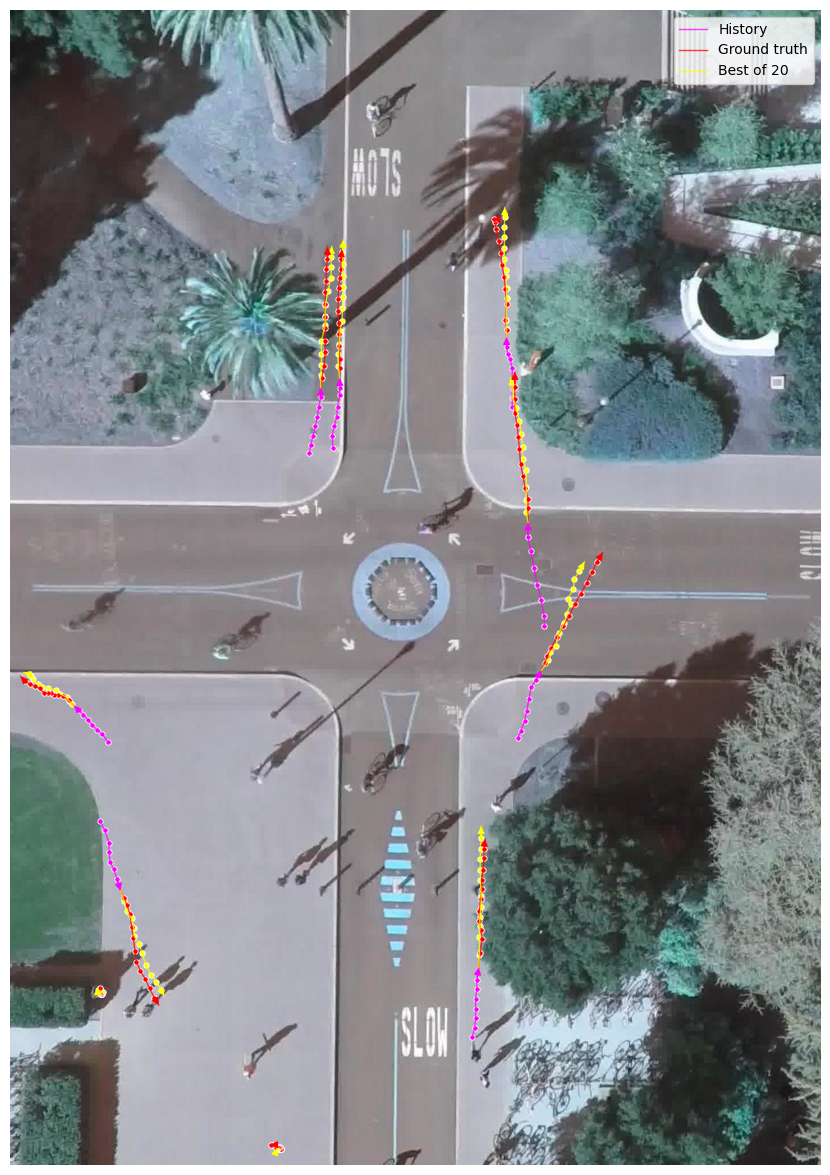}
    \caption{Best of 20 result (SDD)}
    \label{fig:sddbesttraj}
  \end{subfigure}
  \begin{subfigure}[b]{0.25\textwidth}
    \centering
    \includegraphics[width=\textwidth]{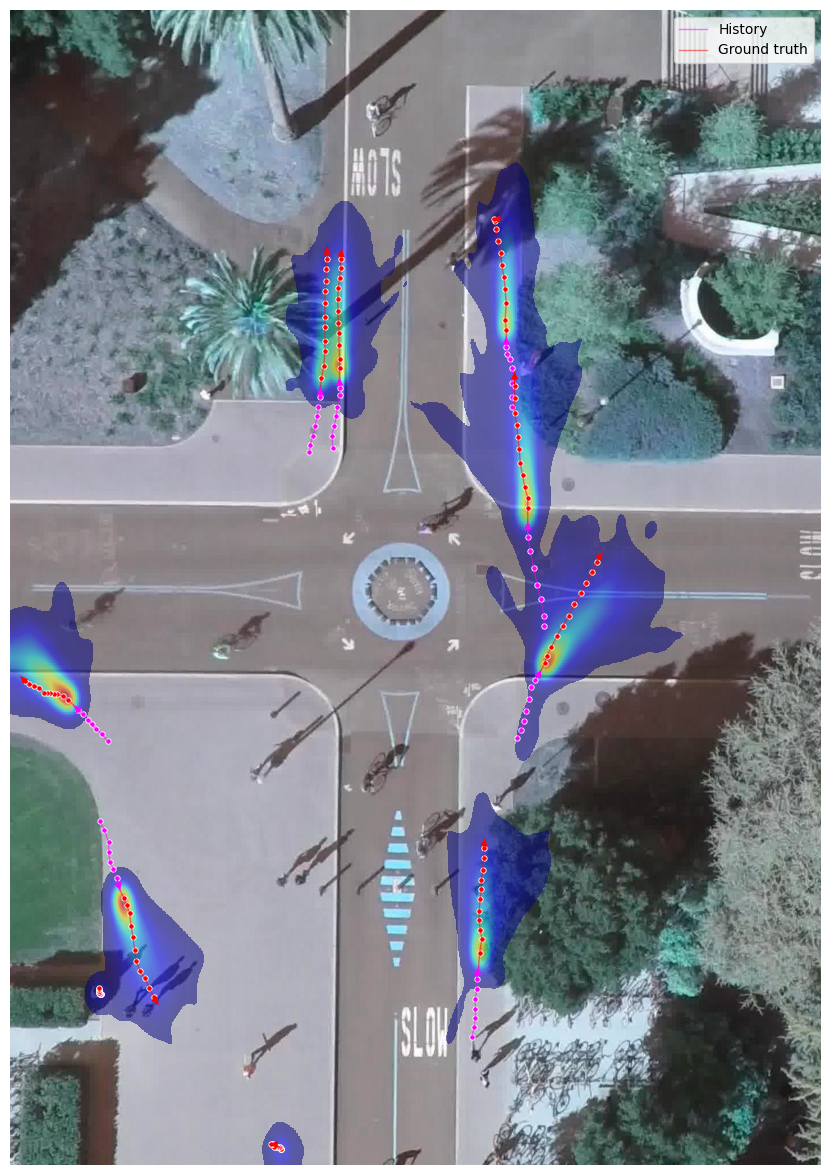}
    \caption{Risk map (SDD)}
    \label{fig:sddriskmap}
  \end{subfigure}
  \begin{subfigure}[b]{0.20\textwidth}
    \begin{minipage}[b]{1.0\textwidth}
      \includegraphics[width=\textwidth]{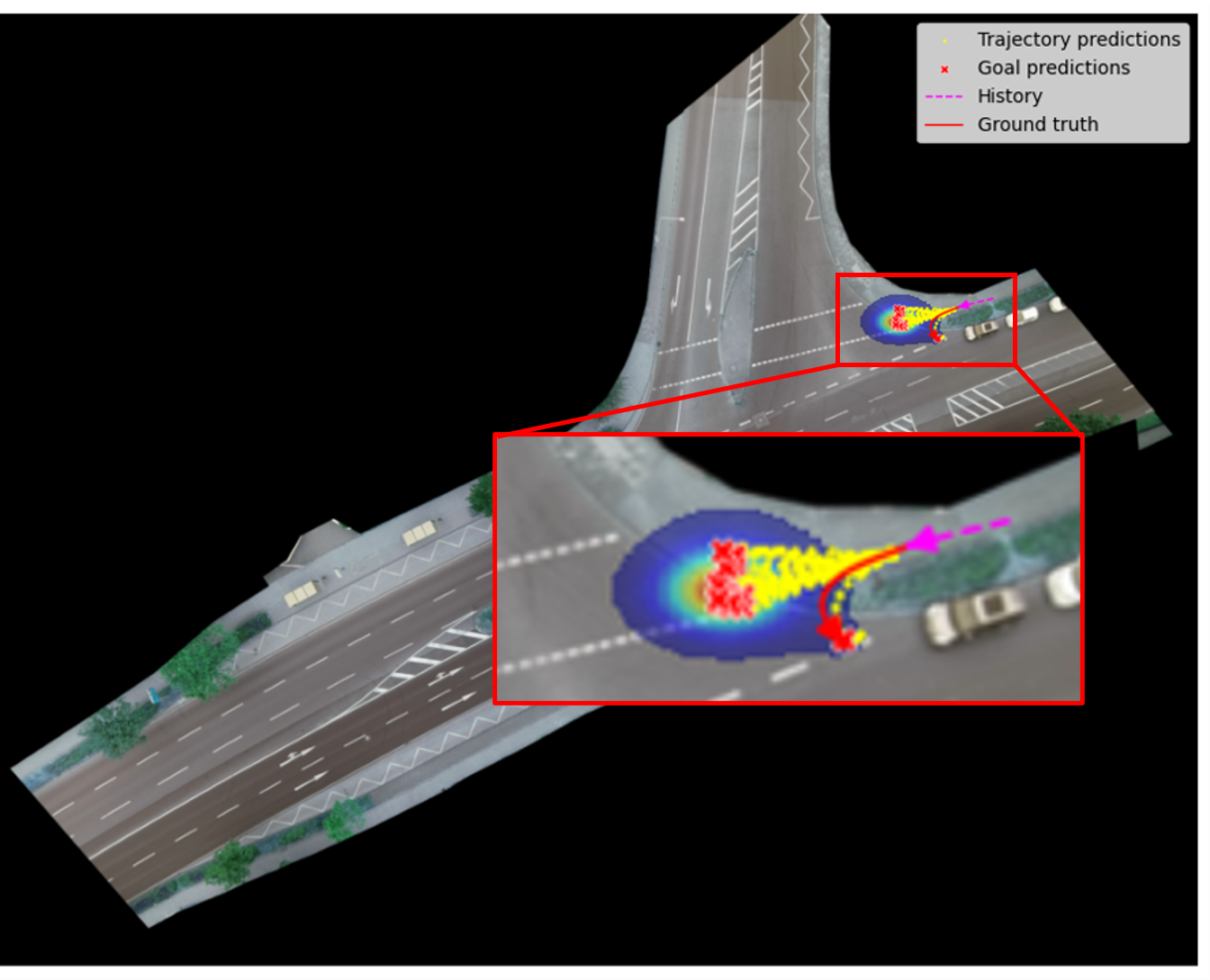}
      \caption{Turning trajectory (inD)}
      \label{fig:straight}
      \includegraphics[width=\textwidth]{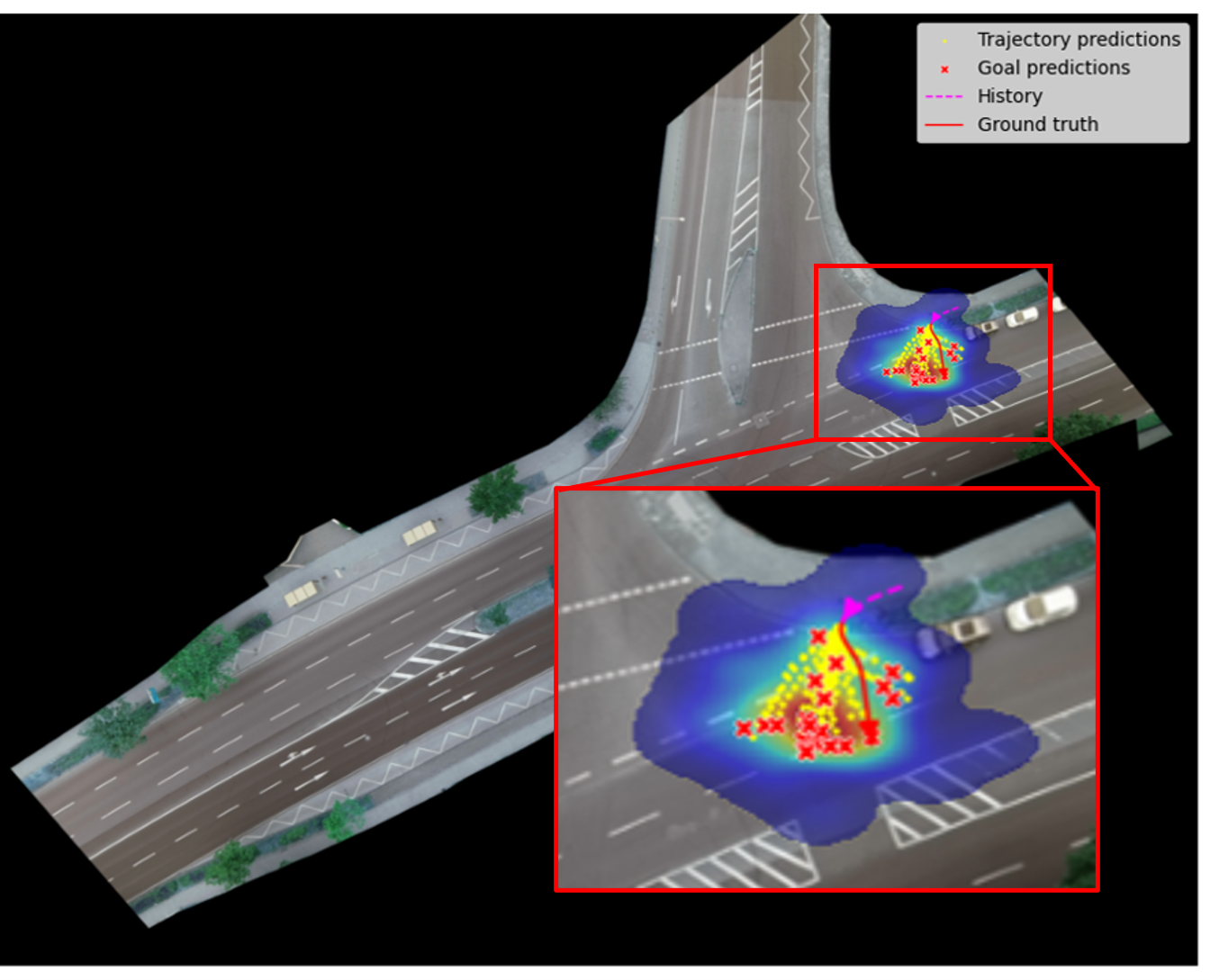}
      \caption{Chaotic prediction (inD)}
      \label{fig:turn}    
    \end{minipage}
  \end{subfigure}
  \caption{Predicted trajectories and risk maps for SDD and inD datasets. Red trajectories are the ground truth, yellow trajectories the prediction and magenta trajectories the partial trajectory used for prediction. The risk map is generated as a function of the scene, with the involvement of all VRUs. Scatter points are sampled goals and trajectories. Best viewed in colour and zoomed in.}
  \label{fig:visualization}
\end{figure*}

Fig. \ref{fig:visualization} presents the qualitative results of our model, depicted here for the inD and SDD datasets. In all presented cases, we see that our model predicts trajectories accurately over time - even in complicated situations like fig. \ref{fig:straight} where a turn in the trajectory is captured and predicted as one of the samples. In contrast, fig. \ref{fig:turn} shows a case of another turning trajectory with a less confident goal sampling, leading to a chaotic trajectory and thus, risk map, prediction. 

\subsection{Ablation Study}

We performed an ablation study, module-by-module, of our proposed model and verified on the inD dataset for its effectiveness in prediction tasks with results shown in Table \ref{table:ablation}. 
Beginning by removing all component modules except a single CVAE layer, trained by the $l_{2}$ loss, produces an FDE and ADE as seen. Continuing, we add modules one after another and compare against the minimum baseline produced by the single CVAE layer. As seen,we notice a consistent decline of ADE and FDE with the addition of each component module thereby showcasing the importance of each of the four enumerated components in our module as well as in pedestrian prediction tasks in general. 

\begin{table}[!h]
    \centering
    \normalsize
    \begin{tabular}{cccc|cc}
    \hline RG & G & N & S & $\text{minADE}_{20}$ & $\text{minFDE}_{20}$ \\
    \hline & & & & 0.58 & 1.31 \\
    $\checkmark$ & & & & 0.35 & 0.65 \\
    $\checkmark$ & $\checkmark$ & & & 0.31 & 0.57 \\
    $\checkmark$ & $\checkmark$ & $\checkmark$ & & 0.31 & 0.56 \\
    $\checkmark$ & $\checkmark$ & $\checkmark$ & $\checkmark$ & \textbf{0.30} & \textbf{0.55} \\
    \hline
    \end{tabular}
    \caption{Ablation studies of our model on the inD dataset. We use a $\text{minADE}_{20}$ and $\text{minFDE}_{20}$ metric for evaluation, with the result reported in meters. Each column represents a component module of the architecture with RG =  Residual GMM, G = Goal Module, N = Neighbour Embedding module, S = Map module}
    \label{table:ablation}
\end{table}
\section{Conclusion} \label{sec:conclusion}

In this paper, we proposed a novel pedestrian trajectory prediction model for stationary BEV scenes in urban areas, based on a Temporal Transformer architecture. In this, we developed a heterogeneous graph attention network to allow for information about the VRU's neighbours' states, the semantic data in map form and the goal position information of the ego agent. Subsequently, we also proposed a novel goal and trajectory generation module using a CVAE-Residual-GMM architecture. We demonstrated the robustness and the generalization capabilities of our model by validating it using three publicly available datasets, in which we beat the state of the art in two, using the same hyperparameters. Future work will focus on the making the model work from the perspective of a moving car and evaluating the performance in real-world scenarios.

A second conclusion we would like to draw from this work is how small the changes in the state of the art has become. We would like to draw attention to Table \ref{table:ind} where the average prediction is smaller than the average stride of a human and while we have beaten the state of the art here, there is a plateauing of scores with changes only in the second decimal. It is our conclusion that we as a community have reached the limits of resolution for prediction in certain situations.





\section*{ACKNOWLEDGMENT}

These researches have been conducted within the AI4CCAM project, financed by the EU Project AI4CCAM with grant number 101076911.  


\clearpage

\bibliographystyle{IEEEtran}
\bibliography{biblio/citations}

\begin{thebibliography}{10}
\providecommand{\url}[1]{#1}
\csname url@samestyle\endcsname
\providecommand{\newblock}{\relax}
\providecommand{\bibinfo}[2]{#2}
\providecommand{\BIBentrySTDinterwordspacing}{\spaceskip=0pt\relax}
\providecommand{\BIBentryALTinterwordstretchfactor}{4}
\providecommand{\BIBentryALTinterwordspacing}{\spaceskip=\fontdimen2\font plus
\BIBentryALTinterwordstretchfactor\fontdimen3\font minus \fontdimen4\font\relax}
\providecommand{\BIBforeignlanguage}[2]{{%
\expandafter\ifx\csname l@#1\endcsname\relax
\typeout{** WARNING: IEEEtran.bst: No hyphenation pattern has been}%
\typeout{** loaded for the language `#1'. Using the pattern for}%
\typeout{** the default language instead.}%
\else
\language=\csname l@#1\endcsname
\fi
#2}}
\providecommand{\BIBdecl}{\relax}
\BIBdecl

\bibitem{helbing1995social}
D.~Helbing and P.~Molnar, ``Social force model for pedestrian dynamics,'' \emph{Physical review E}, vol.~51, no.~5, p. 4282, 1995.

\bibitem{vasishta2017natural}
P.~Vasishta, D.~Vaufreydaz, and A.~Spalanzani, ``Natural vision based method for predicting pedestrian behaviour in urban environments,'' in \emph{2017 IEEE 20th International Conference on Intelligent Transportation Systems (ITSC)}.\hskip 1em plus 0.5em minus 0.4em\relax IEEE, 2017, pp. 1--6.

\bibitem{vasishta2018building}
------, ``Building prior knowledge: A markov based pedestrian prediction model using urban environmental data,'' in \emph{2018 15th International Conference on Control, Automation, Robotics and Vision (ICARCV)}.\hskip 1em plus 0.5em minus 0.4em\relax IEEE, 2018, pp. 247--253.

\bibitem{alahi2016social}
A.~Alahi, K.~Goel, V.~Ramanathan, A.~Robicquet, L.~Fei-Fei, and S.~Savarese, ``Social lstm: Human trajectory prediction in crowded spaces,'' in \emph{Proceedings of the IEEE conference on computer vision and pattern recognition}, 2016, pp. 961--971.

\bibitem{salzmann2020trajectron++}
T.~Salzmann, B.~Ivanovic, P.~Chakravarty, and M.~Pavone, ``Trajectron++: Dynamically-feasible trajectory forecasting with heterogeneous data,'' in \emph{Computer Vision--ECCV 2020: 16th European Conference, Glasgow, UK, August 23--28, 2020, Proceedings, Part XVIII 16}.\hskip 1em plus 0.5em minus 0.4em\relax Springer, 2020, pp. 683--700.

\bibitem{huang2019stgat}
Y.~Huang, H.~Bi, Z.~Li, T.~Mao, and Z.~Wang, ``Stgat: Modeling spatial-temporal interactions for human trajectory prediction,'' in \emph{Proceedings of the IEEE/CVF international conference on computer vision}, 2019, pp. 6272--6281.

\bibitem{gupta2018social}
A.~Gupta, J.~Johnson, L.~Fei-Fei, S.~Savarese, and A.~Alahi, ``Social gan: Socially acceptable trajectories with generative adversarial networks,'' in \emph{Proceedings of the IEEE conference on computer vision and pattern recognition}, 2018, pp. 2255--2264.

\bibitem{lee2017desire}
N.~Lee, W.~Choi, P.~Vernaza, C.~B. Choy, P.~H. Torr, and M.~Chandraker, ``Desire: Distant future prediction in dynamic scenes with interacting agents,'' in \emph{Proceedings of the IEEE conference on computer vision and pattern recognition}, 2017, pp. 336--345.

\bibitem{vemula2018social}
A.~Vemula, K.~Muelling, and J.~Oh, ``Social attention: Modeling attention in human crowds,'' in \emph{2018 IEEE international Conference on Robotics and Automation (ICRA)}.\hskip 1em plus 0.5em minus 0.4em\relax IEEE, 2018, pp. 4601--4607.

\bibitem{mohamed2020social}
A.~Mohamed, K.~Qian, M.~Elhoseiny, and C.~Claudel, ``Social-stgcnn: A social spatio-temporal graph convolutional neural network for human trajectory prediction,'' in \emph{Proceedings of the IEEE/CVF conference on computer vision and pattern recognition}, 2020, pp. 14\,424--14\,432.

\bibitem{mangalam2021goals}
K.~Mangalam, Y.~An, H.~Girase, and J.~Malik, ``From goals, waypoints \& paths to long term human trajectory forecasting,'' in \emph{Proc. International Conference on Computer Vision (ICCV)}, Oct. 2021.

\bibitem{vaswani2017attention}
A.~Vaswani, N.~Shazeer, N.~Parmar, J.~Uszkoreit, L.~Jones, A.~N. Gomez, {\L}.~Kaiser, and I.~Polosukhin, ``Attention is all you need,'' \emph{Advances in neural information processing systems}, vol.~30, 2017.

\bibitem{giuliari2021transformer}
F.~Giuliari, I.~Hasan, M.~Cristani, and F.~Galasso, ``Transformer networks for trajectory forecasting,'' in \emph{2020 25th international conference on pattern recognition (ICPR)}.\hskip 1em plus 0.5em minus 0.4em\relax IEEE, 2021, pp. 10\,335--10\,342.

\bibitem{yuan2021agentformer}
Y.~Yuan, X.~Weng, Y.~Ou, and K.~M. Kitani, ``Agentformer: Agent-aware transformers for socio-temporal multi-agent forecasting,'' in \emph{Proceedings of the IEEE/CVF International Conference on Computer Vision}, 2021, pp. 9813--9823.

\bibitem{yu2020spatio}
C.~Yu, X.~Ma, J.~Ren, H.~Zhao, and S.~Yi, ``Spatio-temporal graph transformer networks for pedestrian trajectory prediction,'' in \emph{Computer Vision--ECCV 2020: 16th European Conference, Glasgow, UK, August 23--28, 2020, Proceedings, Part XII 16}.\hskip 1em plus 0.5em minus 0.4em\relax Springer, 2020, pp. 507--523.

\bibitem{zhang2023forceformer}
W.~Zhang, H.~Cheng, F.~T. Johora, and M.~Sester, ``Forceformer: Exploring social force and transformer for pedestrian trajectory prediction,'' \emph{arXiv preprint arXiv:2302.07583}, 2023.

\bibitem{sadeghian2019sophie}
A.~Sadeghian, V.~Kosaraju, A.~Sadeghian, N.~Hirose, H.~Rezatofighi, and S.~Savarese, ``Sophie: An attentive gan for predicting paths compliant to social and physical constraints,'' in \emph{Proceedings of the IEEE/CVF conference on computer vision and pattern recognition}, 2019, pp. 1349--1358.

\bibitem{xu2022socialvae}
P.~Xu, J.-B. Hayet, and I.~Karamouzas, ``Socialvae: Human trajectory prediction using timewise latents,'' in \emph{European Conference on Computer Vision}.\hskip 1em plus 0.5em minus 0.4em\relax Springer, 2022, pp. 511--528.

\bibitem{xu2023eqmotion}
C.~Xu, R.~T. Tan, Y.~Tan, S.~Chen, Y.~G. Wang, X.~Wang, and Y.~Wang, ``Eqmotion: Equivariant multi-agent motion prediction with invariant interaction reasoning,'' in \emph{Proceedings of the IEEE/CVF Conference on Computer Vision and Pattern Recognition}, 2023, pp. 1410--1420.

\bibitem{mo2022multi}
X.~Mo, Z.~Huang, Y.~Xing, and C.~Lv, ``Multi-agent trajectory prediction with heterogeneous edge-enhanced graph attention network,'' \emph{IEEE Transactions on Intelligent Transportation Systems}, vol.~23, no.~7, pp. 9554--9567, 2022.

\bibitem{chiara2022goal}
L.~F. Chiara, P.~Coscia, S.~Das, S.~Calderara, R.~Cucchiara, and L.~Ballan, ``Goal-driven self-attentive recurrent networks for trajectory prediction,'' in \emph{Proceedings of the IEEE/CVF Conference on Computer Vision and Pattern Recognition}, 2022, pp. 2518--2527.

\bibitem{messaoud2021trajectory}
K.~Messaoud, N.~Deo, M.~M. Trivedi, and F.~Nashashibi, ``Trajectory prediction for autonomous driving based on multi-head attention with joint agent-map representation,'' in \emph{2021 IEEE Intelligent Vehicles Symposium (IV)}.\hskip 1em plus 0.5em minus 0.4em\relax IEEE, 2021, pp. 165--170.

\bibitem{mangalam2020not}
K.~Mangalam, H.~Girase, S.~Agarwal, K.-H. Lee, E.~Adeli, J.~Malik, and A.~Gaidon, ``It is not the journey but the destination: Endpoint conditioned trajectory prediction,'' in \emph{Computer Vision--ECCV 2020: 16th European Conference, Glasgow, UK, August 23--28, 2020, Proceedings, Part II 16}.\hskip 1em plus 0.5em minus 0.4em\relax Springer, 2020, pp. 759--776.

\bibitem{wang2022stepwise}
C.~Wang, Y.~Wang, M.~Xu, and D.~J. Crandall, ``Stepwise goal-driven networks for trajectory prediction,'' \emph{IEEE Robotics and Automation Letters}, vol.~7, no.~2, pp. 2716--2723, 2022.

\bibitem{yue2022human}
J.~Yue, D.~Manocha, and H.~Wang, ``Human trajectory prediction via neural social physics,'' in \emph{European Conference on Computer Vision}.\hskip 1em plus 0.5em minus 0.4em\relax Springer, 2022, pp. 376--394.

\bibitem{huang2023multimodal}
R.~Huang, H.~Xue, M.~Pagnucco, F.~Salim, and Y.~Song, ``Multimodal trajectory prediction: A survey,'' \emph{arXiv preprint arXiv:2302.10463}, 2023.

\bibitem{kothari2021interpretable}
P.~Kothari, B.~Sifringer, and A.~Alahi, ``Interpretable social anchors for human trajectory forecasting in crowds,'' in \emph{Proceedings of the IEEE/CVF Conference on Computer Vision and Pattern Recognition}, 2021, pp. 15\,556--15\,566.

\bibitem{deo2020trajectory}
N.~Deo and M.~M. Trivedi, ``Trajectory forecasts in unknown environments conditioned on grid-based plans,'' \emph{arXiv preprint arXiv:2001.00735}, 2020.

\bibitem{guo2022end}
K.~Guo, W.~Liu, and J.~Pan, ``End-to-end trajectory distribution prediction based on occupancy grid maps,'' in \emph{Proceedings of the IEEE/CVF Conference on Computer Vision and Pattern Recognition}, 2022, pp. 2242--2251.

\bibitem{jia2022hdgt}
X.~Jia, P.~Wu, L.~Chen, H.~Li, Y.~Liu, and J.~Yan, ``Hdgt: Heterogeneous driving graph transformer for multi-agent trajectory prediction via scene encoding,'' \emph{arXiv preprint arXiv:2205.09753}, 2022.

\bibitem{ronneberger2015unet}
O.~Ronneberger, P.~Fischer, and T.~Brox, ``U-net: Convolutional networks for biomedical image segmentation,'' 2015.

\bibitem{robicquet2016learning}
A.~Robicquet, A.~Sadeghian, A.~Alahi, and S.~Savarese, ``Learning social etiquette: Human trajectory understanding in crowded scenes,'' in \emph{Computer Vision--ECCV 2016: 14th European Conference, Amsterdam, The Netherlands, October 11-14, 2016, Proceedings, Part VIII 14}.\hskip 1em plus 0.5em minus 0.4em\relax Springer, 2016, pp. 549--565.

\bibitem{inDdataset}
J.~Bock, R.~Krajewski, T.~Moers, S.~Runde, L.~Vater, and L.~Eckstein, ``The ind dataset: A drone dataset of naturalistic road user trajectories at german intersections,'' in \emph{2020 IEEE Intelligent Vehicles Symposium (IV)}, 2019, pp. 1929--1934.

\bibitem{breuer2020opendd}
A.~Breuer, J.-A. Term{\"o}hlen, S.~Homoceanu, and T.~Fingscheidt, ``opendd: A large-scale roundabout drone dataset,'' in \emph{2020 IEEE 23rd International Conference on Intelligent Transportation Systems (ITSC)}.\hskip 1em plus 0.5em minus 0.4em\relax IEEE, 2020, pp. 1--6.

\bibitem{wang2019heterogeneous}
X.~Wang, H.~Ji, C.~Shi, B.~Wang, Y.~Ye, P.~Cui, and P.~S. Yu, ``Heterogeneous graph attention network,'' in \emph{The world wide web conference}, 2019, pp. 2022--2032.

\bibitem{sohn2015learning}
K.~Sohn, H.~Lee, and X.~Yan, ``Learning structured output representation using deep conditional generative models,'' \emph{Advances in neural information processing systems}, vol.~28, 2015.

\bibitem{dempster1977maximum}
A.~P. Dempster, N.~M. Laird, and D.~B. Rubin, ``Maximum likelihood from incomplete data via the em algorithm,'' \emph{Journal of the royal statistical society: series B (methodological)}, vol.~39, no.~1, pp. 1--22, 1977.

\bibitem{pellegrini2009you}
S.~Pellegrini, A.~Ess, K.~Schindler, and L.~Van~Gool, ``You'll never walk alone: Modeling social behavior for multi-target tracking,'' in \emph{2009 IEEE 12th international conference on computer vision}.\hskip 1em plus 0.5em minus 0.4em\relax IEEE, 2009, pp. 261--268.

\bibitem{lerner2007crowds}
A.~Lerner, Y.~Chrysanthou, and D.~Lischinski, ``Crowds by example,'' in \emph{Computer graphics forum}, vol.~26, no.~3.\hskip 1em plus 0.5em minus 0.4em\relax Wiley Online Library, 2007, pp. 655--664.

\bibitem{bertugli2021ac}
A.~Bertugli, S.~Calderara, P.~Coscia, L.~Ballan, and R.~Cucchiara, ``Ac-vrnn: Attentive conditional-vrnn for multi-future trajectory prediction,'' \emph{Computer Vision and Image Understanding}, vol. 210, p. 103245, 2021.

\bibitem{sadeghian2018trajnet}
A.~Sadeghian, V.~Kosaraju, A.~Gupta, S.~Savarese, and A.~Alahi, ``Trajnet: Towards a benchmark for human trajectory prediction,'' \emph{arXiv preprint}, 2018.

\bibitem{amirian2020opentraj}
J.~Amirian, B.~Zhang, F.~V. Castro, J.~J. Baldelomar, J.-B. Hayet, and J.~Pettr{\'e}, ``Opentraj: Assessing prediction complexity in human trajectories datasets,'' in \emph{Proceedings of the asian conference on computer vision}, 2020.

\bibitem{liang2020garden}
J.~Liang, L.~Jiang, K.~Murphy, T.~Yu, and A.~Hauptmann, ``The garden of forking paths: Towards multi-future trajectory prediction,'' in \emph{Proceedings of the IEEE/CVF Conference on Computer Vision and Pattern Recognition}, 2020, pp. 10\,508--10\,518.

\end{thebibliography}

\end{document}